\documentclass{article} 
\usepackage{nips14submit_e,times}
\usepackage{hyperref}
\usepackage{url}
\usepackage{wrapfig}

\title{Why (and When and How) Contrastive Divergence Works}
\author{Ian E. Fellows\\
Fellows Statistics \\
\texttt{http://www.fellstat.com}\\
\texttt{ian@fellstat.com}
}
\usepackage{amsmath} 
\usepackage{amssymb}
\usepackage{natbib}
\usepackage[belowskip=-5pt,aboveskip=0pt]{caption}
\usepackage{graphicx}
\usepackage[margin=1.5in]{geometry}
\DeclareMathOperator*{\argmin}{\arg\!\min}
\DeclareMathOperator*{\argmax}{\arg\!\max}

\newcommand{\KL}{\rm{KL}}

\nipsfinalcopy 

\begin{document}
   \maketitle
\begin{abstract}
Contrastive divergence (CD) is a promising method of inference in high dimensional distributions with intractable normalizing constants, however, the theoretical foundations justifying its use are somewhat weak. This document proposes a framework for understanding CD inference, including how and when it works. It provides multiple justifications for the CD moment conditions, including framing them as a variational approximation. Algorithms for performing inference are discussed and are applied to social network data using an exponential-family random graph models (ERGM). The framework also provides guidance about how to construct MCMC kernels providing good CD inference, which turn out to be quite different from those used typically to provide fast global mixing.
\end{abstract}
\section{Introduction to Contrastive Divergence}

Performing inference in the high dimensional data setting presents unique computational challenges as traditional inference methods such as maximum likelihood quickly become intractable. An alternative inference procedure is contrastive divergence (CD) \cite{hinton_2002}, which has the advantage of being relatively easy to compute stochastically resulting in it becoming a widely used technique in the context of Boltzmann machines \citep{fischer_2014}. This paper will focus on providing a theoretical foundation for CD, with particular attention devoted to the case of exponential-family models.

Let $Y$ be a multi-dimensional random variable with realization $y$, $p(Y)$ be the true distribution of the data generating process, and $q(Y)$ be a family of of distributions that we are attempting to fit to the data. 
For example, $q$ may be defined as being from an exponential-family
$$
q(Y) = \frac{1}{z(\eta)}e^{\eta g(Y) +o(Y)}~~~y\in{\cal Y},
$$
 where $\eta$ is a vector of parameters, 
i$g$ is a vector of sufficient statistics, $o$ is an offset statistic, and $z$ is the normalizing constant defined as $z(\eta) = \sum_{y'in{\cal Y}}e^{\eta g(y') + o(y')}$\cite{bar78}. As the dimensionality of $y$ increases, the sum in $z$ quickly becomes intractable to compute exactly (except in special cases), thus the actual numeric value of $q$ is difficult to evaluate. For every possible $\eta$, the distribution may also be parameterized by a unique and equivalent set of mean value parameters $\mu_\eta = E_q(g(Y))$\cite{bar78}.
 
Suppose we have one observation $y$ drawn from the true distribution $p$. It is well known that maximum likelihood inference is equivalent to minimizing the Kullbeck-Liebler (\KL) divergence between $p$ and $q$ with the KL divergence is defined as $\KL(p || q) = \sum_y \log(\frac{p(y)}{q(y)})p(y)$. Minimizing the divergence yields
$
\KL(p||q) \propto -E_{p}(log(q(Y))) \approx -\log(q(y)),
$
which is the sample negative log-likelihood. 

Unfortunately, in many cases it is impossible to evaluate the log-likelihood directly because it contains an intractable normalizing constant. As a solution to this problem CD introduces a related distribution $T_q(Y | y'),~y'\in{\cal Y})$, which is defined to be a valid Gibbs transition kernel for $q$ (i.e. $q(y) = \sum_{y^{(0)}}T_q(y |y^{(0)})q(y^{(0)}) = E_{q}(T_q(y|Y))$. The contrastive divergence objective function is defined as
\begin{equation}\label{eq:cdobj}
\KL(p || q) - \KL(T_q(Y|y) || q).
\end{equation}

The first term in the objective function is identical to that of maximum likelihood inference. The second is the deviation between the Markov chain distribution starting at the observed data vector $y$ and $q$. Typically augmented objective functions of this form are justified by observing that when $p=q$ the objective function is minimized, or nearly minimized \citep{lyu_2011}. Clearly this is true with the first term of the equation, but it is not necessarily true for the second. When $p=q$ the objective function becomes $-\KL(T_p(Y|y^{(0)}=y) || p)$, implying that the function reaches its minimum at $p$ if $T$ and $q$ are maximally divergent at that point in the neighborhood of $p$. The second term can be made to be arbitrarily close to 0 by choosing $T_q$ to be the result of many steps of an MCMC process. Specifically, suppose $T_q^k$ represents the transition probability resulting from applying k steps of an MCMC process, then \citet{cover_2006} show that
$
\KL(T_q^k||q) \leq \KL(T_q^{k-1}||q),
$
with convergence assured under regularity conditions. This assures that provided the chain length of CD is sufficiently large, the method becomes equivalent to maximum likelihood. Unfortunately, even though the convergence of MCMC is exponential, it may take an extraordinarily long time to reach the equilibrium distribution. In many common implementations a chain length of 1 is used, resulting in a distribution that is starkly different from $q$.

If $q$ belongs to an exponential family, then the gradient of the objective function is
\begin{equation}\label{eq:cdgrad}
\frac{\delta}{\delta \eta_j}(\KL(p || q) - \KL(T_q || q)) = -E_p(g_j(Y)) + E_{T_q}(g_j(Y)) - \frac{\delta T_q}{\delta \eta_j}\frac{\delta \KL(T_q || q)}{\delta T_q}.
\end{equation}
The first term on the RHS of equation \ref{eq:cdgrad} can be approximated using the observed values from the data ($g_j(y)$), and the second expectation can be approximated by sampling from $T_q$ which is relatively inexpensive computationally. The third term is problematic in that it can not be estimated without evaluating $q$. \citet{hinton_2002} suggested dropping it based on simulations that suggested that it ``is small and seldom opposes the resultant of the other two terms.'' This then results in the contrastive divergence moment conditions
\begin{equation}\label{eq:cdmom}
g(y) = E_{T_q}(g(Y)).
\end{equation}
There is, however, no theoretical reason to believe that the third term is ignorable in general. In the context of Restricted-Boltzmann machines, the bias of the gradient approximation in equation \ref{eq:cdmom} relative to the maximum likelihood gradient goes to 0 as the number of MCMC steps in $T_q$ goes to infinity \cite{bengio_2009}, \cite{fischer_2011}. However, this result relies upon the chain $T_q$ approaching $q$ as the number of steps $k$ increases, which is unrealistic as the typical number of steps is very small.

\section{Reframing the Problem}

There are two main theoretical problems with the current development of CD. Firstly, it is questionable whether the objective function in equation \ref{eq:cdobj} finds a $q$ close to $p$ when it is minimized,. Secondly, it is unknown whether the minimization is adequately achieved when ignoring the third term of equation \ref{eq:cdgrad}. In order to address these issues, we develop an alternative objective function which provides a more principled foundation for the CD gradient approximation. First we will introduce some notation. Let $y$ be indexed as $y_i$ for $i \in (1,...,m)$,  $b$ be a vector of sets of indices of $y$, with union $a = \bigcup_i b_i$. Let $r(b)$ represent the probability of selecting indices $b$. We can then define the joint probability of an observed Gibbs chain as
$
T_q(Y^{(1:k)},B | y^{(0)}) = \prod_{i=1}^k q(Y^{(i)}_{B_i} | Y_{\backslash B_i}^{(i-1)})r_i(B_i),
$
where $B_i$ are the indices of $Y$ which have non zero probability of changing at step $i$.

The conditional probability of the last step of the MCMC process, given $A$ is
$$
q^*(Y^{(k)} | y^{(0)},a) = \frac{1}{\pi(a)} \sum_{a = \bigcup_ib_i}\sum_{Y^{(1:k-1)}} T_q(Y^{(1:k)}, b | y^{(0)}) 
$$
where $\pi(A) = \sum_{A  = \bigcup_ib_i}\prod_i^k r(b_i)$. Because the Gibbs update is a valid transition kernel for the unconditional distribution, $q^*$ is a valid kernel for the conditional distribution given $a$ in that it satisfies detailed balance.
\begin{eqnarray}
\sum_{y^{(0)}_a} q^*(Y^{(k)} | y^{(0)},a) q(y^{(0)}_a | y^{(0)}_{\backslash a}) &=& \frac{1}{\pi(a)} \sum_{a = \bigcup_ib_i}(\prod_i^k r(b_i))\sum_{y^{(0)}_a} q(y^{(0)}_a | y^{(0)}_{\backslash a})\sum_{Y^{(1:k-1)}} \prod_{i=1}^k q(Y^{(i)}_{B_i} | Y_{\backslash B_i}^{(i-1)})  \nonumber \\
&=&\frac{1}{\pi(a)} \sum_{a = \bigcup_ib_i}(\prod_i^k r(b_i))q(Y^{(k)}_a | Y^{(k)}_{\backslash a} = y^{(0)}_{\backslash a})   \nonumber \\
&=& q(Y^{(k)}_a | Y^{(k)}_{\backslash a} = y^{(0)}_{\backslash a}) \nonumber
\end{eqnarray}
Thus, while the MCMC distribution $T_q$ is unlikely to be close to the equilibrium distribution $q$ after a small number of steps, if the MCMC sampler is chosen appropriately $q^*$ may be very close to $q(Y_{a} | y_{\backslash a})$ because of the drastically reduced dimensionally (i.e. $|a| << m$).

\subsection{Augmented Divergence}

Now let us consider an augmented objective function for a single subset $a$. We define this to be
\begin{eqnarray}
d_a(p,q) &=& \KL(p||q) - \KL(p_m||q_m) + \sum_y\log(\frac{q(Y_{a} | y_{\backslash a})}{q^*(Y_{a}| y_{\backslash a},y^{(0)})})p(y) \\
&=& \KL(p || q^*p_m) \geq 0 \nonumber
\end{eqnarray}
where $p_m(y_{\backslash a}) = \sum_{y_a}p(y)$ and $q_m(y_{\backslash a}) = \sum_{y_a}q(y)$ are the marginal distributions of $p$ and $q$. Similarly, let us define the conditional distributions as $p_c = p(Y_{a} | y_{\backslash a})$ and $q_c = q(Y_{a} | y_{\backslash a})$. 

To see why $d_a$ is a good augmentation let us consider the first two terms. From \citet{huber_1985} and \citet{lyu_2011} we know that $\KL(p||q) - \KL(p_m||q_m) \geq 0$, with equality when $p=q$. The third term represents the discrepancy between the conditional distribution of $q$ and the approximation $q^*$ averaged over the true distribution, and when $p=q$ it reduces to 
$
d_A(p,p) = \KL(p || p^*p_m) \approx 0. 
$
Thus, $d$ reduced to near its lower bound when $q$ fits the data generating distribution perfectly. Further, if $p^* = p_c$ then the equality is achieved. This equation can also be used to justify setting $y^{(0)}$ to be from the sample data, as starting the MCMC chain from the equilibrium distribution $p$ will cause (at least approximately) the resulting chain $q^*$ to be closest to $q_c$ when $q=p$. This implies that even if the lower bound is not achieved at $p=q$, the minimum will be nearly achieved, and the divergence may be approximated with
\begin{eqnarray}
\argmin_q d_a(p,q) &=& \argmin_{q^*} \KL(p || q^*p_m)  \nonumber \\
&=& \argmax_{q^*} E_p(\log(q^*(Y_{a_i}| y_{\backslash a},y^0))) \nonumber \\
&\approx& \argmax_{q^*}  \log(q^*(y_{a_i}| y^0)) \\ \nonumber
&=& \argmax_{q^*} \ell_a(q^* | y).
\end{eqnarray}
So the augmented divergence is approximately minimized at by maximizing the observed log likelihood of the transitional distribution $q^*$. 

\subsection{Combined Augmented Divergence}

$d_a$ is a good measure of the local information about $p$ within $a$, but we wish to have a more global criterion. Now we will combine all possible subsets to generate a global augmented divergence by simply taking a weighted average.
\begin{equation} \label{eq:gibbs}
cd(p,q) = \sum_A d_{A}(p,q)\pi(A) = E_\pi(\ell_A(q^* | y)).
\end{equation}
Since each one of the individual $d_{A}$ obtains a minimum at or near 0 when $p \approx q$, the combined augmented divergence $cd$ is also a reasonable measure of the divergence between $p$ and $q$ aggregating the local information in each subset $A$. Small $cd$ values close to 0 indicate good agreement between $p$ and $q$ over the subsets, and large values indicate poor model fit. We may then minimize $cd$ subject to $q$ in order to find a good model fit.

If $q$ is exponential family, and $q^* \approx q_c$ (which is not unreasonable even for modest numbers of MCMC steps), then taking the gradient and setting it to zero yields
\begin{eqnarray}
0 &=& E_\pi(\frac{\delta}{\delta q}\log(q^*(y_a| y)) \nonumber \\
&\approx& E_\pi( g(y) - E_{q^*(Y_A|y_{\backslash A})}(g(Y))) \nonumber \\
&=& g(y) - E_T(g(Y)). \nonumber
\end{eqnarray}
Thus, provided that our MCMC approximation is reasonable within the restricted sample space of $a$, the minimum of $cd$ occurs approximately when the contrastive divergence moment conditions $g(y) = E_T(g(Y))$ are satisfied. 

If $q$ is not exponential family or if $q^*$ is appreciably different from $q$, then we may still have some hope that the moment conditions will provide good inferences. The quantity in equation \ref{eq:gibbs} can be recognized as the empirical analog of a set of generalized method of moments conditions $E_{T_q}(E_\pi(\frac{\delta}{\delta q}\log(q^*(Y_A| y^{(0)}))))=0$. This implies that minimizing $cd$ is equivalent to fitting $T$ via moment equations. In general it is not true that these moment equations reduce to the CD moment conditions, but using the CD moment conditions will provide a fit of $T$ which matches the data and thus will likely drive $cd$ to near its minimum.

\subsection{A Variational Approximation to $\ell_A$}
Even if $q^*$ is not close to $q_c$ in distribution, the contrastive divergence moment conditions may be derived as a variational approximation to $cd$. Let $q$ belong to an exponential family. We may approximate a single augmented divergence with:
\begin{eqnarray}
\log(q^*(y^{(k)} | a)) &=& \sum_{b : a = \bigcup_ib_i}\sum_{y^{(1:(k-1))}}(\log(\frac{q^*(y^{(1:k)},b|a)}{f(y^{(1:(k-1))},b|a)}) f(y^{(1:(k-1))},b|a) \nonumber \\
 &&+ \log(\frac{f(y^{(1:(k-1))},b|a)}{q^*(y^{(1:(k-1))},b | y^{(k)},a)}) f(y^{(1:(k-1))},b|a)) \nonumber \\
 &\gtrsim & \sum_{b : a = \bigcup_ib_i}\sum_{y^{(1:(k-1))}}\log(\frac{q^*(y^{(1:k)},b|a)}{f(y^{(1:(k-1))},b|a)}) f(y^{(1:(k-1))},b|a) \label{eq:v1}\\
 &=& E_f(log(q_c(y^{(k)}_{B_k} | Y^{(k-1)}_{\backslash B_k}))) \nonumber \\
 &&- \KL(f(y^{(1:k-1)},b|a) || q^*(y^{(1:k-1)},b|a)) \label{eq:v2}\\
 &=& U(y^{(k)},q,f) \nonumber
\end{eqnarray}
where $f$ is an arbitrary distribution, and with equality in \ref{eq:v1} obtained when $q^*(y^{(1:(k-1))}, b | Y^{(k)}, a) = f(y^{(1:(k-1))},b|a)$. The derivative of equation \ref{eq:v2} is then
\small
$$
\frac{\delta U(y^{(k)},q,f)}{\delta \eta}= g(y^{(k)}) - E_f(E_{q_c}(g(Y^{(k)}) | Y^{(k-1)}_{\backslash B_k})) + \sum_{i=1}^{k-1} E_f(g(Y^{(i)})) - E_f(E_{q_c}(g(Y^{(i)}) | Y^{(i-1)}_{\backslash B_i}))  
$$
\normalsize
Substituting $f = q^*(y^{(1:k-1)},b | a)$ into the derivative causes the terms in the summation to cancel out yielding
$$
\frac{\delta U(y^{(k)},q,f=q^*)}{\delta \eta} = g(y^{(k)}) - E_{q^*}(g(Y^{(k)}) | a).
$$ Note that the inequality in \ref{eq:v1} is tight as $\KL(q^*(y^{(1:(k-1))},b | Y^{(k)},a) || q^*(y^{(1:(k-1))},b | a))$ is small, especially when $k$ is large. We may combine the individual results by minimizing equation \ref{eq:gibbs}, resulting in
$
E_\pi(\log(q^*(y^{(k)} | a)))  \gtrsim  E_\pi(U(y^{(k)},q,f = q^*)).
$
Taking taking the derivative, and setting it to 0 gives us the fixed point
$
g(y^{(k)}) = E_{T_q}(g(Y^{(k)})),
$
which we recognize as the contrastive divergence moment conditions.

So we have three principled ways of justifying our use of the contrastive divergence moment conditions within the combined augmented divergence framework. First, if $q^*$ is close enough to $q_c$ to be approximately an exponential family, then the moment conditions fall rather naturally out of $cd$. Otherwise, by observing that minimizing $cd$ is equivalent to finding a GMM fit of $T$, the CD moment conditions can be justified as an alternate GMM fit. Further, the conditions may be derived from a variational approximation to $cd$.

\section{Special Cases}\label{sec:ex}

For most cases, the solution to equation \ref{eq:gibbs} is not tractable analytically, requiring us to use the CD moment conditions to find an approximate solution. There are however some cases where we can solve the equation directly, and these examples will provide us some intuition about when CD is likely to work well, and when it has limited utility.

\textbf{Composite Likelihood:} Suppose that $T$ is a single step of a blocked Gibbsian kernel such that
$
T(Y,A | y^{(0)}) = q_c(Y_{A} | y^{(0)}_{\backslash A})r(A).
$
We can see that in this case, our $q^*$ simplifies to $q^*(Y^{(k)} | y^{(0)},a) = q(Y_{A} | y^{(0)}_{\backslash A})$, and if we let $y^{(0)} = y$, then the objective function becomes
$$
\argmax_q \sum_A \log(q_c(Y_{A}=y_A | y_{\backslash A})) r(A),
$$
which we recognize as the objective function for composite likelihood. If $q$ is exponential family, the first derivative is
$
\frac{\delta}{\delta \eta}= g(y) - E_{T}(g(Y))
$
with a second derivative of
$
\frac{\delta^2}{\delta \eta_i \delta \eta_j}= cov_{T}(g_i(Y),g_j(Y)).
$
So in this case, the CD moment conditions are identical to the moment conditions for composite likelihood.

If $r$ is chosen to be the uniform distribution over $A$ such that $A$ only contains a single element, then the objective function becomes
$
\argmax_q \sum_{i=1}^m\log(q(Y_{i}=y_i | y_{\backslash i})),
$
which is known as the pseudo-likelihood. Pseudo-Likelihood is a simple form of composite likelihood inference and will serve as a baseline to evaluate CD inference in section \ref{sec:exp}. 

The more elements present in $A$, the closer the composite likelihood estimate will be to the maximum likelihood estimate. On the other hand, the computational complexity of sampling from $q_c$ increases exponentially as more indices are added. Alternatively, but relatedly, suppose that $T$ selects a subset $A$ with probability $\pi(A)$, and then runs an MCMC sampler within $A$ on the conditional distribution $q_c$ until the chain reaches equilibrium. Then the transition kernel can be written as
$$
T(Y^{(k)},A | y^{(0)}) = \pi(A)\sum_{i=1}^{k-1}\prod_{i=1}^k t_i(Y^{(i)}_{B_i} | Y^{(i-1)},B_i.y^{(0)})r_i(B_i|Y^{(i-1)})
= q_c(Y_{A}=y_A | y^{(0)}_{\backslash A})\pi(A)
$$
which by identical argument results in a composite likelihood solution with weights defined by $\pi$. So, if $T$ results in the exact equality of $q^*$ and $q_c$, either though direct sampling via a single blocked Gibbs step, or though running a sampler to convergence within each $A$, then the CD objective function is equivalent to the composite likelihood objective function.

\textbf{Single Scan Gibbs:}
Suppose that $T$ is a Gibbsian kernel sequential updating $y_s:y_{s+k}$ with $k\leq m$, $s$ chosen at random, and indices wrapping when needed. Then the log likelihood for a given $A$ is
$$
 \ell_A(q^* | y) = \log(q^*(y^i_{A}| y^i_{\backslash A},y^{(0)}_{A}))  
 = \sum_{i=1}^k \log(q(Y_i=y_i | y_{\backslash A}, y_{s:s + i-1},y^0_{(i+1):(s+k)})) 
$$
and setting our initial MCMC state equal to our observation (i.e. $y^{(0)}=y$) we arrive at
$
 \ell_A(q^* | y) = \sum_{i=1}^k \log(q(Y_i=y_i | y^{(0)}_{\backslash i}).
$
Applying equation \ref{eq:gibbs} results in
$
\argmax_q \sum_{i=1}^m\log(q(Y_{i}=y_i | y_{\backslash i})),
$
which is identical to the pseudo-likelihood function, indicating that our longer chain gave no additional improvement over a single step of random scan Gibbs. Indeed, any k-step Gibb update which makes only one pass though $A$ will result in a pseudo-likelihood objective function, so if we wish to make non-pseudo-likelihood inferences, the kernel must revisit indices that it has already sampled.

\textbf{Conditional Independence:}
Suppose that $T$ is a single step blocked Gibbs kernel. Let $r$ be such that it chooses $A$ to be the set of two indices at random $i,j$ such that they are conditionally independent (i.e. $q(Y_i,Y_j | y_{\backslash \{i,j\}}) = q(Y_i | y_{\backslash \{i,j\}})q(Y_j | y_{\backslash \{i,j\}})$). With $y^0=y$ we obtain
$
 \ell_A(q^* | y) =  \log(q(Y_i=y_i | y_{\backslash i})q(Y_j=y_j | y_{\backslash j})),
$
and minimizing equation \ref{eq:gibbs} results in the objective function
$
\argmax_q \sum_{i=1}^m\log(q(Y_{i}=y_i | y_{\backslash i})) \sum_{j=1}^mr(\{i,j\}),
$
which is again the objective function for pseudo-likelihood (weighted by $\sum_{j=1}^mr(\{i,j\})$), so if non-pseudo-likelihood inferences are the goal, then $A$ must be chosen such that there is dependence among its members.

\textbf{One and a Half Pass Gibbs:}
Suppose that $q$ represents an exponential family, and that $m=2$. Let $T$ be a Gibbs sampler sampling $y_1$ then $y_2$, then back to $y_1$. In this case, with $y^0=y$ we obtain
$$
 \ell_A(q^* | y) =   \log(q(Y_1=y_1 | y_{\backslash 1})\sum_{y_1'}q(Y_1=y_1' | y_{\backslash 1})q(Y_2=y_2 | y_{\backslash \{1,2\}},y_1=y_1')).
$$
Taking the derivative yields
\small
$$
\frac{\delta  \ell_A(q^* | y)}{\delta \eta_i} = g_i(y) - E_{Y_1 | y_{\backslash 1}}(g_i(Y)) + 2E_{Y' | y_{\backslash 1}}(g(Y')f - g(Y'))- E_{Y' | y_{\backslash 1}}(fE_{Y_2 | Y_{\backslash 2}'}(g(\{Y_1',Y_2,y_{3:m}\})))
$$
\normalsize
where $f = \frac{q(Y_2=y_2 | Y_1=y_1')}{E(q(Y_2=y_2 | Y_1=Y_1'))}$, and $Y'=\{Y_1',y_{2}\}$. Each of the expectations in the derivative may be approximated by sampling from $T$, however it is apparent that even with this simple Gibbsian updating, the likelihood becomes difficult to express. More complex $q^*$ representing longer MCMC chains will only increase this complexity. Thus it is useful (even for the above simple case) to use the CD moment conditions to approximately optimize the objective.

\section{Learning Algorithms}\label{sec:learn}

In the context of maximum likelihood inference in exponential families, one can perform maximum likelihood inference using the stochastic approximation algorithm \citep{robbins_1951}, where the update equation
$$
\eta^{i+1} = \eta^i + \gamma_i (g(y) - E_q(g(Y)|\eta^i))
$$
converges to the maximum likelihood solution when the gradient coefficient is allowed to decrease geometrically (i.e. $\gamma_i = \frac{a}{i}$ for some $a>0$). The challenge here from a computation perspective is that though $E_q$ can be theoretically estimated by MCMC, the chain length required for convergence is too long to make this practical. This is precisely the reason that the CD moment conditions are attractive, as they do not require full convergence of the MCMC chain, rather only a limited number of steps ($k$).

The CD update step takes the simplistic approach of replacing $q$ in the stochastic approximation with $T$ yielding
$$
\eta^{i+1} = \eta^i + \gamma_i (g(y) - E_T(g(Y)|\eta^i)).
$$
$T$ is very easy and quick to sample from, since each sample takes just a few steps from an MCMC process to compute. The expectation can then be approximated by simply taking the sample expectation ($\frac{1}{n}\sum_i^ng(y^i)$ where $y^i$ are independent samples from $T$). \citet{yuille_2004} showed that this update equation converges under some regularity conditions, and provided some conditions under which the converged upon value is the maximum likelihood solution. Unfortunately, the conditions required for convergence to the maximum likelihood are unlikely to be met unless the MCMC chain is run long enough to reach equilibrium.

Stochastic approximation approaches are similar to gradient descent, in that convergence is generally linear in the number of iterations. There may be some advantage to be gained by using a  Newton-like update. If $q^*$ is sufficiently close to $q_c$ to be approximately exponential family, then the second derivative within equation \ref{eq:gibbs} is 
$
H(\eta)= cov_{T}(g(Y),g(Y)),
$
which is identical to the composite likelihood hessian in section \ref{sec:ex}. We can then use a Newton-like update
$$
\eta^{i+1} = \eta^i - [H(\eta_i)]^{-1} (g(y) - E_T(g(Y)|\eta^i)).
$$
Both the hessian and expectation can be approximated with the same set of samples from $T$, so this algorithm takes similar computational effort to the stochastic gradient algorithm at each step. If the hessian is a good approximation, this algorithm will see quadratic convergence to the CD solution. The down side is that convergence is not guaranteed, especially when the hessian approximation is poor.

\section{Experiments}\label{sec:exp}

One popular class of exponential families are Exponential-family Random Graph Models (ERGM) \citep{fra86}. These models have been used widely to investigate the structure of social interactions \citep{hunter_2008}, and are typically fit either by finding the maximum pseudo-likelihood estimate (MPLE) \citep{dujin_2009} or using MCMC methods to find a maximum likelihood estimate (MCMC-MLE) \citep{hunter_2006}. As such, they provide a useful case study for approximate inferential methods which have less computational complexity than MCMC-MLE and more accurate inference than MPLE. ERGMs model the presence or absence of a relationship between a set of nodes which typically represent individual people. Let the dyad $y_{i,j}$ be 1 if there is a connection between node $i$ and $j$, and 0 if there is none.  Connections are considered to be undirected (i.e. $y_{i,j}=y_{j,i}$).

The dataset we use comes from the National Longitudinal Study of Adolescent Health (Add Health), which in addition to collecting data on health related behaviors, also recorded information about the social networks of subjects \citep{Harris_2003}. One particular high-school surveyed by the study had 1270 students, with 361, 309, 346, and 254 students in 9th, 10th, 11th and 12th grade respectively. Students were asked to select up to five close male and five close female students as friends. A relationship is considered to be present if and only if both students nominate each other. We will consider a simple four parameter ERGM. The first term in the model is the number of edges in the network (${\rm edges} = \sum_{i>j}y_{i,j}$). The second term is the number of students with no connections (${\rm isolates} = \sum_{i} I(\sum_jy_{i,j} >0)$, where $I$ is the indicator function). We expect that more connections are present between members of the same grade, so the third term will model this with the count of connections between students of the same grade (${\rm nodeMatch}= \sum_{i>j} I(grade_i = grade_j)y_{i,j}$). Another important social relationship is transitivity, which is the tendency of the friend of my friend to also be my friend. Simplistic statistics used to model this have the tendency to display poor statistical properties such as phase transitions\cite{han03}. For this reason, a more robust measure known as the geometrically weighted edgewise shared partner (GWESP) statistic is used. See \citet{hunter_2006} for the mathematical specification of this statistic as well as a more thorough justification. GWESP statistics also require the specification of a curve parameter $\alpha$, which we set to value of $\frac{2}{3}$ based on the MLE goodness of fit. Finally, since students can only select up to a maximum of 10 friends, we should incorporate this constraint into the model. This is done by including an offset which is set to $-\infty$ if any node has more than 10 connections, and 0 otherwise.

The special cases outlined in section \ref{sec:ex} lead us to believe that the choice of MCMC algorithm will be important to improving our estimate over the MPLE. The dimensionality of $y$ is 805,815, so a sequential scan Gibbian update would require a huge number of steps before it revisits any elements of $y$. Similarly we might expect a random scan Gibbsian update to perform similarly due to the low probability of revisiting dyads. For this reason, we consider three alternate MCMC schemes based on the intuition that dyads connected to the same node are unlikely to be conditionally independent. The Node-$s$ MCMC kernel proceeds by selecting (for each chain) a node at random, and then selecting $s$ dyads incident to that node. A simple random scan Gibbs is then performed for $k$ steps within the selected dayds. 

\begin{figure}
\includegraphics[scale=0.65]{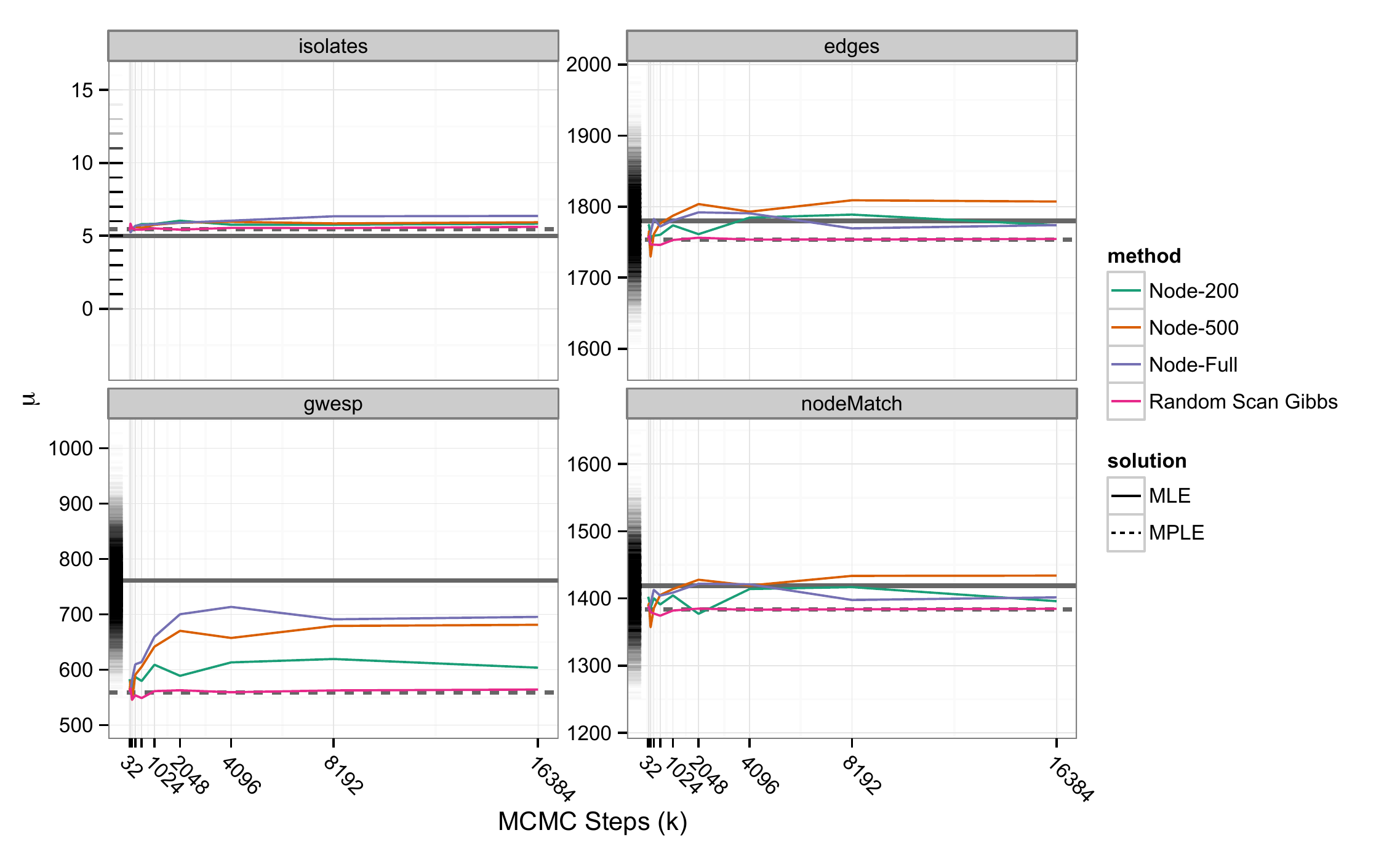}
\caption{\label{fig:f1} Mean value parameter estimates for different MCMC methods and step lengths. The distributions on the left edges of the plots represent the distributions of the statistics $g(Y)$ at the MLE.}
\end{figure}

Figure \ref{fig:f1} shows the mean value parameters for CD inference using Node-$s$ kernels with $s=$ 200, 500 and 1269 (Full) at different chain lengths up to $2^{14}$. Also shown are the results for random scan Gibbs, and the MLE and MPLE estimates. Each CD fit was done using the Newton-like updates in section \ref{sec:learn}, leading to much faster convergence than the gradient update. Mean values $\mu$ are shown rather than natural parameters $\eta$ because $\mu$ is more useful in determining how well a parameter set matches the MLE. The pseudo-likelihood estimate fits well for the isolates, edges and nodeMatch terms as the mean values are all within one standard deviation of the MLE's mean value among networks simulated from the MLE model. The MPLE grossly underestimates the amount of transitivity in the network as represented by the GWESP term, with a mean value more than 3 standard deviations lower than the MLE. The distribution of the GWESP term under the MLE is displayed in the density band along the left side of the plot, which shows that the average GWESP term for networks generated under the MPLE model are virtually unheard of in networks generated by the MLE. This indicates that there is significant room for improvement on this fit.

The Random Scan Gibbs solutions are indistinguishable from pseudo-likelihood, even with a $k$ of 16,384. This confirms our theoretical result that the asymptotics of $T$ approaching $q$ as $k$ increases provides a poor foundation for CD-k inference, as a hugely massive $k$ would be required in order to reach equilibrium in the high dimensional space of $y$. However, by restricting our Gibbs sampling to be within subsets of $y$, significant improvement can be made over pseudo-likelihood. For the GWESP term, we see that as $s$ and $k$ increase, the solutions become closer to the MLE, though there still exists some downward bias. We also see some improvement in the edges and nodeMatch terms relative the the MPLE, and all estimates are close to the MLE for the isolates parameter. Node-Full reaches near its best estimate when $k \geq 2048$, which is less than 2 times larger than the dimensionality of the subset (1269); Far too few to reach full equilibrium, but enough so that elements of $y$ have a high probability of being revisited.

\section{Conclusion: Why, When and How}

This paper explored \textbf{why} contrastive divergence works using a novel combined augmented divergence framework. We illustrated how this new $cd$ objective function alleviates some of the theoretical concerns with the original objective function and provided three justifications of the CD moment conditions. From the examples in section \ref{sec:ex} we may surmise that MCMC kernels work for CD \textbf{when} they visit and revisit highly related sections of $y$ in a modest number of steps, otherwise CD inference will have no advantage over the Pseudo-Likelihood solution. Finally, we showed \textbf{how} to implement CD using a Newton-like update equation. The simulation study conducted used the Newton-like update with success, and validated our heuristics about what types of kernels are best.

   \bibliographystyle {unsrt}
\bibliography {contrastive}    

\end{document}